\documentclass{isrpaper}

\usepackage{orcidlink}
\usepackage{amsmath,amssymb,amsfonts}
\usepackage{graphicx}
\usepackage{textcomp}
\usepackage{booktabs}
\usepackage{tabularx}
\usepackage{xcolor}

\usepackage[
    backend=biber,
    style=authoryear,
    natbib=true,
    doi=false,
    url=false,
    isbn=false,
    maxbibnames=99,
    maxcitenames=2,
    uniquelist=false,
    giveninits=true,    
    terseinits=true,    
]{biblatex}

\AtEveryBibitem{%
   \clearfield{Title}%
   \clearfield{eprint}%
   \clearfield{note}%
}

\renewcommand{\cite}[1]{\unskip\nolinebreak[3]\ \parencite{#1}}

\usepackage{etoolbox}
\AtBeginEnvironment{tabular}{\footnotesize}
\AtBeginEnvironment{tabularx}{\footnotesize}
\AtBeginEnvironment{tabular*}{\footnotesize}
\usepackage[font=small,labelfont=bf,labelsep=period,
            justification=justified,format=plain,singlelinecheck=false]{caption}
\title{Calibrated Uncertainty for Trustworthy Clinical Gait Analysis Using Probabilistic Multiview Markerless Motion Capture}

\author{
Seth Donahue\,\orcidlink{0000-0002-8387-9887}\,$^{1,2,*}$,
Irina Djuraskovic\,\orcidlink{0009-0008-8940-6790}\,$^{3,4,*}$,
Kunal Shah\,$^{3}$,
Fabian Sinz\,$^{5,6}$,
Ross Chafetz\,$^{7}$,
R. James Cotton\,\orcidlink{0000-0001-5714-1400}\,$^{3,4}$}

\affiliations{\footnotesize
$^{1}$Shriners Children's Lexington, Lexington, KY;\;
$^{2}$Dept. of Physical Therapy, University of Kentucky, Lexington, KY;\;
$^{3}$Shirley Ryan AbilityLab, Chicago, IL;\;
$^{4}$Northwestern University, Evanston, IL;\;
$^{5}$University of G\"ottingen, G\"ottingen, Germany;\;
$^{6}$Lower Saxony Center for AI \& Causal Methods in Medicine;\;
$^{7}$Shriners Children's Philadelphia, Philadelphia, PA.\;
$^{*}$Indicates co-first authors.}

\paperdate{December 2025}

\correspondence{rcotton@sralab.org}

\paperabstract{\textbf{Objective:} This study evaluates the calibration and accuracy of a probabilistic MMMC method. Video-based human movement analysis holds potential for movement assessment in clinical practice and research. However, the clinical implementation and trust of multi-view markerless motion capture (MMMC) require that, in addition to being accurate, these systems produce reliable confidence intervals to indicate how accurate they are for any individual. Building on our prior work utilizing variational inference to estimate biomechanical variables against clinical gold-standards. \textbf{Methods:} We analyzed data from 68 participants across two institutions, validating the model against an instrumented walkway and standard marker-based motion capture. We measured the calibration of the confidence intervals using the Expected Calibration Error (ECE). \textbf{Results:} The model demonstrated reliable calibration, yielding ECE values generally $< 0.1$ for both step and stride length and bias-corrected gait kinematics. We observed a median step and stride length error of $\sim 16$ mm and $\sim 12$ mm respectively, with median bias-corrected kinematic errors ranging from $1.5^{\circ}$ to $3.8^{\circ}$ across lower extremity joints. Consistent with the calibrated ECE, the magnitude of the model's predicted uncertainty correlated strongly with observed error measures. \textbf{Conclusion:} These findings indicate that, as designed, the probabilistic model reconstruction quantifies epistemic uncertainty, shown by low absolute error and calibrated uncertainty. \textbf{Significance:} These findings highlight the potential to identify unreliable outputs without the need for concurrent ground-truth instrumentation through the use of this probabilistic model.\\[4pt]
}

\begin{document}
\maketitle

\section{Introduction}
\label{sec:introduction}
Accurate and accessible human movement analysis through computer-vision-based methods holds the potential to democratize clinical gait analysis beyond the motion laboratory. Recent years have seen considerable progress in the assessment of computer-vision-based human movement analysis for clinical purposes. Current multiview markerless motion capture  (MMMC) models present accurate aggregate performance compared to external validation measures\cite{pierzchlewicz_optimizing_2023, kanko_concurrent_2021, antognini_reframe_2025, dsouza_comparison_2024, Uhlrich2023_OpenCap}. However, they lack quantifiable uncertainty outputs necessary for the end user to know when the data is trustworthy, which is a significant barrier for clinical implementation. Calibrated uncertainty estimates are essential for detecting and excluding instances of low-quality biomechanical reconstruction, thereby improving the reliability and trust of kinematic data used for clinical decision-making.

Our prior work established a method to provide statistically sound confidence intervals, for kinematic estimates from MMMC through a probabilistic model. It demonstrated strong internal validity, compared to the most reliable keypoints used as the pseudo-ground truth \cite{cotton_confidence_2025}. However, the probabilistic model has not undergone external validation against clinical systems \cite{cotton_confidence_2025}. The present work fills this gap by evaluating the calibration and external validity of a probabilistic MMMC framework.

\subsection{Computer Vision and Differentiable Models}
Recent advances in computer vision, paired with GPU-accelerated physics engines such as MuJoCo \cite{todorov_mujoco_2012} and musculoskeletal models optimized for MuJoCo like MyoSuite \cite{caggiano_myosuite_2022}, have substantially improved the accuracy and feasibility of estimating human motion directly from video \cite{cotton_differentiable_2024}.  Differentiable pipelines involve the joint optimization of a differentiable biomechanical model using an implicit function that maps from time to the joint angle trajectory. The poses are then optimized end-to-end to minimize reprojection error between the predicted marker locations and the detected keypoints. These approaches outperform traditional multistage pose estimation pipelines that rely on sequential 2D detection, triangulation, smoothing, and the inverse kinematics from these virtual keypoint trajectories \cite{cotton_optimizing_2023, cotton_improved_2023, cotton_differentiable_2024,unger_2024_differentiable}.

End-to-end differentiable MMMC leverages two complementary structural constraints. First, geometric
redundancy across cameras provides a principled mechanism to down-weight inconsistent keypoints. Second, biomechanical priors—such as fixed segment lengths, joint coupling, and anatomically feasible ranges of motion—restrict the solution space to physically plausible motions. Recent work has demonstrated that such models can achieve high accuracy for clinical tasks \cite{unger_2024_differentiable, cotton_differentiable_2024}.

However, despite their strong overall performance, previous MMMC systems can also sometimes produce outliers or noisy results, such as in instances where the number of views are reduced due to suboptimal camera geometry, occlusions or space constraints such as in a hallway. This can lead to 3D systematic bias due to miscalibration or issues in the triangulation of the keypoints respectively. The quality and anatomical accuracy of the detected keypoints is critical for the downstream kinematic analysis, but their fidelity is often limited by the use of sparser keypoint datasets, different marker conventions and anatomical landmarks between Human Pose Estimation algorithms and mis-association of the keypoints in instances where a patient may be assisted by a therapist.

Without trustworthy confidence intervals accompanying their output, end-users either require comparison of each trial against gold-standard measurements or designing \emph{ad hoc} heuristics for quality control. Aggregated accuracy metrics, typically reported in most validation studies, can mask systematic reductions in accuracy across variations in anatomy, assistive device use, or pathological gait patterns. As MMC moves from research to clinical deployment, it is critical to incorporate robust and validated confidence intervals to ensure reliability across diverse patient populations and clinical environments.

\subsection{Probabilistic Modeling, Validation and Calibration}
\begin{figure*}[!tbp]
\centering
\includegraphics[width=1\linewidth]{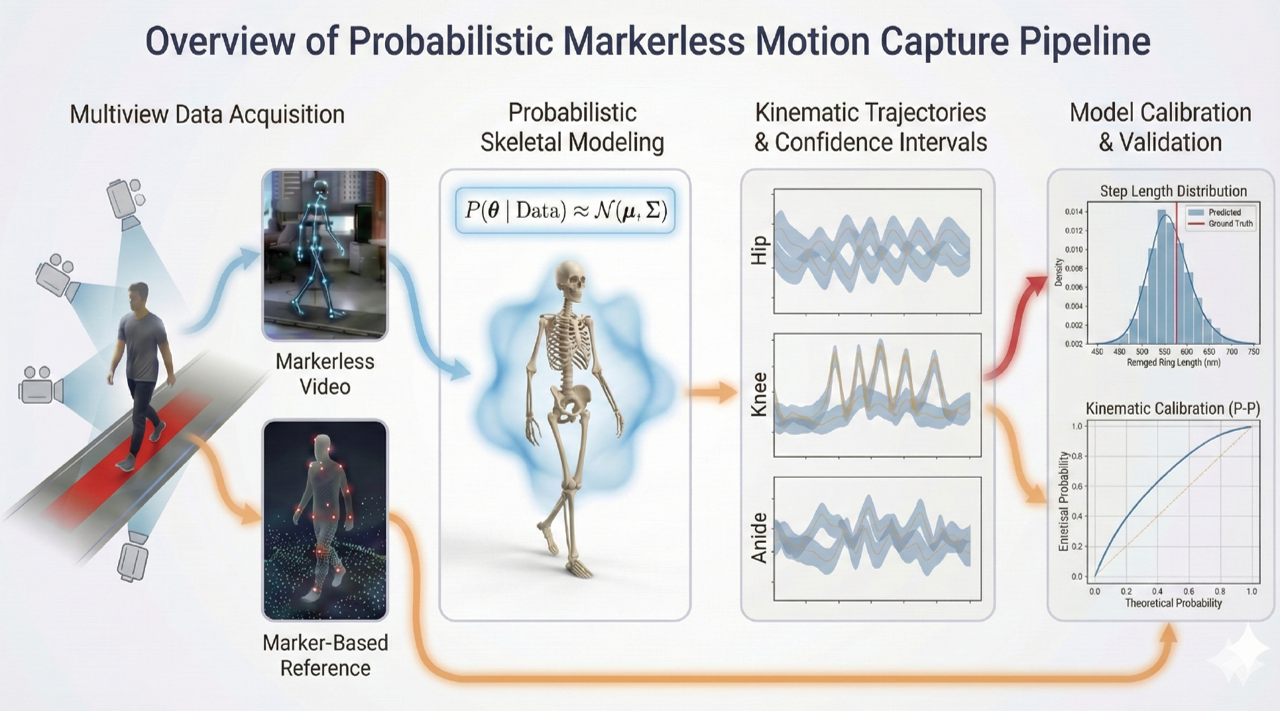}
\caption{Overview of the Probabilistic MMMC Validation Pipeline. The pipeline proceeds in four stages: (1) Multiview Data Acquisition collects synchronized video from varying camera configurations alongside either marker-based or GaitRite references across diverse cohorts, including pediatric, neurologic, and prosthetic users. (2) Probabilistic Modeling estimates pose parameters, outputting joint angles and uncertainty in the measures. (3) Outputs from the probabilistic model are the expected (mean of the posterior) trajectory and associated per-time and per-joint standard deviation, which can be presented as a 95\% CI. (4) Model Calibration \& Validation evaluates the error between systems for the biomechanical variables and the distribution and reliability of the output estimated uncertainties. (Figure generated with assistance of Google Nano Banana)}
\label{fig:probabilistic_overview}

\end{figure*}

In recent work, we demonstrated that an implicit trajectory representation (i.e., a neural network that maps from time to joint angles) can be extended to a fully probabilistic formulation and fit to multiview data using variational inference\cite{cotton_confidence_2025}. This addresses two coupled challenges: first, most keypoint detectors are not inherently probabilistic and second, lack a well-defined likelihood function to reflect the uncertainty of predicted keypoint locations. Our formulation learns this likelihood jointly during inference and propagates uncertainty through the full pipeline to output a confidence interval over kinematic estimates given multiple noisy observations of virtual marker locations captured from different cameras. (Fig. \ref{fig:probabilistic_overview}).

Kinematic MMMC methods often report their validity through the calculation of average errors against ground-truth measurements across multiple trials using statistical tools such as Analysis of Variance, or Statistical Parametric Mapping. However,  while such methods validate that the method is accurate on average,  we have found those metrics to be limited as they do not tell us specifically \emph{when} we can trust the output as it does not provide an estimate of the uncertainty of the measurements at every point in time. For example occlusions or camera geometry in one session or trial may make it unreliable compared to the average performance. Knowing when the output can be trusted is crucial for healthcare and high-risk settings such as surgical decision making \cite{Huang2020_Calibration}.

To asses and ensure MMMC-data quality (video only data), we propose the use of a probabilistic model that outputs both biomechanical trajectories and the associated uncertainties at every time-point. Therefore, for the probabilistic model to be validated, it  should satisfy two conditions:  that the error between systems for the biomechanical variables is small (traditional validation), and the distribution of the output uncertainties are well-calibrated (probabilistic validation). Both can be quantified with comparison against gold-standard clinical systems. Specifically, calibration refers to how closely a nominal predicted (X\%) confidence interval contains X\% of the clinically measured data  \cite{gneiting_probabilistic_2007}. We quantify global probabilistic calibration with the Expected Calibration Error (ECE), which measures the discrepancy between the model’s predicted confidence interval coverage and the coverage observed against ground-truth measurements.  A low ECE indicates the model’s uncertainty estimates are aligned with the error between the model and the gold standard measurements.

\subsection{Purpose and key contributions}
The purpose of this study is to evaluate whether the kinematic reconstructions and associated confidence intervals produced by our probabilistic MMMC framework are both accurate and well calibrated when assessed against external, clinically validated measurement systems. We evaluated model calibration across different populations and camera configurations and validated performance using two datasets from two sites.  One compared step length measured against an instrumented walkway (GaitRite), and another compared joint kinematics from simultaneously measured marker-based kinematics.

The present work makes the following key contributions:
\begin{itemize}

 \item We demonstrate that the probabilistic model yields accurate spatial gait variables (stride length error less than the average Minimal Detectable Change (MDC) for the GaitRite). We further show that the uncertainty (i.e., gait parameter confidence intervals) of the model is well calibrated with respect to the instrumented walkway with small ECE values. We also show that both the accuracy and calibration of the probabilistic model holds across multiple different clinical populations.

 \item We show that kinematic reconstructions are accurate relative to clinically accepted marker-based motion capture outputs with small errors up to a bias, $<5^{\circ}$, which is the Minimal Clinically Important Difference (MCID). In addition, we demonstrate that the uncertainty of the kinematic trajectories are well calibrated with respect to marker based motion capture.  This requires applying a participant-wise bias correction to account for systematic differences between the biomechanical model in our clinical pipeline versus the probabilistic methods.

\item We show that confidence metrics produced by a well-calibrated model have the potential to be used to automatically identify and remove unreliable steps, producing a workflow for trustworthy data analysis in larger-scale clinical practice.

 \end{itemize}

\section{Methods}

\subsection{Shirley Ryan Participants}
The Northwestern University Institutional Review Board (IRB) approved this study. We collected data from a diverse population of individuals seen for rehabilitation at the Shirley Ryan AbilityLab. The dataset consists of  416 trials over 41 ($N = 41$) participants. The population included able-bodied individuals ($n = 9$) and persons with gait impairments from various etiologies, including lower-limb prosthesis users ($n = 10$), persons with neurologic gait impairments such as stroke, traumatic brain injury, or spinal cord injury ($n = 18$), as well as a pediatric cohort ($n = 4$). The adult population had an average age of $46.73\pm 19.0$ years, an average height of $170.0 \pm 11.9$ cm, and an average weight of $77.5 \pm 24.8$ kg. The pediatric population had an average age of $13.3\pm 3.6$ years, an average height of $155.9 \pm 22.1$ cm, and an average weight of $54.7 \pm 18.0$ kg. The data was collected as individuals walked across an instrumented walkway (GaitRite).  The trials consisted of a combination of walking at varying speeds, using an orthosis or different assistive devices, sometimes with physical assistance from therapists.  In the neurologic group, assistive devices were used in 62 of the 170 trials, including a regular cane, quad cane, crutch, or rolling walker. In the pediatric group, a quad cane was used in 4 of the 20 trials.

\subsection{Shriners Participants}
An external IRB for Shriners Children’s (Protocol \#PHL2305) approved this study. Each participant, or their legal guardian in the case of minors, provided informed consent prior to participation. Individuals with a range of clinical diagnoses underwent comprehensive gait analysis. The sub-set of a larger study included a convenience sample of 27 participants with 133 walking trials with an average age of $11.3 \pm 3.3$ years, an average height of $137.7 \pm 20.6$ cm, and an average weight of $45.0 \pm 26.4$ kg. The cohort presented with a diverse set of primary diagnoses, most commonly Cerebral Palsy ($n=12$), several other orthopedic and neurological conditions. Participants used a variety of assistive devices, including forearm crutches and posterior rolling walkers, though the majority ($n=24$) required no assistance.

\subsection{Marker-based Data Processing}
Three-dimensional lower-extremity kinematics were collected using a 12-camera Vicon Vantage motion capture system (Vicon Motion Systems Ltd., Oxford, UK) operating at $60\,\text{Hz}$. Reflective markers were placed by trained, certified physical therapists according to the Shriners Children’s Gait Model (SCGM) template \cite{kruger_shriners_2024} (or a variant using a thigh marker in place of the patellar marker). Clinical anthropometric data (e.g., leg length, hip/knee/ankle width) were entered into the SCGM model in Vicon Nexus (v2.15.0). Each session included a static calibration and a minimum of three barefoot walking trials performed with the participant's customary assistive device, if required. Marker trajectories were processed using standard SCGM procedures: labeled, gap-filled, and low-pass filtered (fourth-order Butterworth, $10\,\text{Hz}$), with joint kinematics computed in Python 3.11.

\subsection{Markerless Data Collection and Preprocessing}
Video data from both collection sites were collected and preprocessed as described in \cite{cotton_differentiable_2024}. In brief,  video data from the Shirley Ryan AbilityLab were recorded with synchronized multiview video cameras acquired using our custom software \cite{Multi_Camera_Repo} from 8 to 12 FLIR BlackFly S GigE (Teledyne Vision Solutions, Waterloo, ON, Canada) cameras recorded at $30\,\text{Hz}$. Video data from the Shriners Children's utilized synchronized calibrated multi-camera video collected in Vicon Nexus using eight FLIR BlackFly S USB3 cameras recorded at $60\,\text{Hz}$.

Processing of both datasets was as follows: images were processed with the MetrABs-ACAE algorithm \cite{sarandi_learning_2023}, which was trained on the superset of keypoints, and we retained the 87 keypoints from the MOVI dataset \cite{ghorbani_movi_2021}. Our prior work found these keypoints worked well for biomechanics, with the dense coverage of the whole body stabilizing estimates of the trunk and torso \cite{cotton_optimizing_2023, cotton_improved_2023}. Easymocap was used for scene reconstruction and segmentation \cite{dong2021fast}, and a custom visualization tool was used to annotate the participant \cite{cotton_differentiable_2024}.

\subsection{Markerless Probabilistic Calculations and Hyperparameter Optimization}

The probabilistic model is a MuJoCo based, GPU-accelerated and differentiable biomechanical model from which we obtain our mean and covariance probability distribution for each joint and time-point. Time was scaled to remain below $\pi$ and was positionally encoded using an encoding dimension that would ensure a minimum frequency of 80Hz. For detailed discussion and derivation of the probabilistic model's derivation, optimization, implementation, and internal validation against a deterministic model see \cite{cotton_confidence_2025}.  Our goal is to compute the posterior distribution of joint angles given a set of detected keypoints and their confidence scores, which in theory can be computed with Bayes' rule:
\begin{align*}
p(\boldsymbol \theta_t \mid \mathbf Y_t, \mathbf S_t) = \frac{p(\mathbf Y_t, \mathbf S_t \mid \boldsymbol \theta_t) \, p(\boldsymbol \theta_t)}{p(\mathbf Y_t, \mathbf S_t)},
\end{align*}

where $\boldsymbol \theta_t \in \mathbb R^{40}$ are the kinematic joint angles at time $t$, and $(\mathbf Y_t, \mathbf S_t) = \left\{ (\mathbf y_t^c  \in \mathbb R^{87 \times 2}, \mathbf s_t^c  \in \mathbb R^{87}) \, \forall \, c \right\}$ represents the set of all keypoints and their confidence scores across $C$ cameras.  This formulation is unsolvable for two reasons, first:  $p(\mathbf Y_t, \mathbf S_t)$ is intractable due to the high dimensionality of $\boldsymbol \theta_t$; second: the likelihood $p(\mathbf Y_t, \mathbf S_t \mid \boldsymbol \theta_t)$ was not directly available. Therefore, in order to solve this, we have adopted a variational inference approach, approximating the posterior with a Gaussian distribution of joint angles $q_\phi(\boldsymbol \theta_t)$ and minimizing the Kullback-Leibler (KL) divergence to the true (intractable) posterior. The gaussian exponential form was selected based on previous work, as it out-preformed other likelihood functions for the reconstruction and estimation of uncertainty parameters \cite{cotton_confidence_2025}

Following \cite{cotton_confidence_2025}, the distribution $q_\phi$ is parameterized by an implicit function $f_\phi$ that maps time to the trajectory distribution, parameterized by both the mean and covariance of the trajectory:
\begin{align*}
f_\phi: t \mapsto (\boldsymbol \mu_\phi(t), \mathbf u_\phi(t)), \quad
q_\phi(\boldsymbol \theta_t) = \mathcal N(\boldsymbol \theta_t; \boldsymbol \mu_\phi(t), \Sigma_\phi(t)),
\end{align*}
where $\boldsymbol \mu_\phi \in \mathbb R^{40}$ and $\mathbf u_\phi \in \mathbb R^{40 \times R}$ parameterize a low-rank representation of the covariance $\Sigma_\phi(t)$.  At each time point, the model predicts a mean and standard deviation (uncertainty) estimate of the kinematic data.
The posterior also models the correlation structure over joint angles, capturing correlations between lumbar bending and hip flexion, as well as between numerous other intuitively expected correlations \cite{cotton_confidence_2025}.

To fit the probabilistic model to the video based data, we optimized a composite objective loss (Eq~\ref{eq:loss}) that combines probabilistic reconstruction, geometric consistency, and calibration accuracy.  While the core objective is based on the Evidence Lower Bound (ELBO), we augmented the loss with specific regularization terms to enforce physically valid trajectories and well-calibrated keypoint uncertainties:

\begin{align}
\mathcal{L} &= \text{NLL}
- \lambda_{\theta}\,H(\theta)
+ \lambda_{\text{site}} \,\lVert \Delta_{\text{site}} \rVert^{2} \nonumber \\[2mm]
&\quad + \lambda_{\text{excess}} \,\mathcal{L}_{\text{excess}}
+ \lambda_{\text{ece}} \, (\text{ECE} \cdot N_{\text{kp}}), \label{eq:loss}
\end{align}

The primary terms of the loss function were the negative log-likelihood (NLL) of the observed 2D keypoints and the entropy of the variational pose distribution $H(\theta)$, which together form the standard ELBO objective. Geometric regularizers included $\lambda_{\text{site}} \lVert \Delta_{\text{site}} \rVert^{2}$, to minimize site-offset errors, and $\mathcal{L}_{\text{excess}}$, to penalize joint-limit violations. These joint limits within anatomical bounds are enforced by passing the network's outputs through a tanh nonlinearity and then rescaling them to match those specified by the model. A pose prior penalizes samples that exceed the specified joint limits during variational optimization. It is from the Normal joint angle distributions output from the probabilistic model that we use as a basis to extract meaningful biomechanical data. The mean of the joint angle distribution is what we use for the calculation of reconstruction errors. Spatial measures, step and stride length, we sampled from the distributions of the joint angles, and then input each sample into a forward kinematics model from which we can extract marker positions and calculate either step or stride length.

ECE in Eq~\ref{eq:loss} is an internal calibration estimate computed by treating highest-confidence detected keypoints as ground truth detections and measuring the consistency of those keypoints relative to the estimated probability distribution. While this estimate is not perfect, as we do not have ground truth data from video alone, it plays an important role in estimating the noise from the detected keypoints. See \cite{cotton_confidence_2025} for a detailed derivation. The hyperparameter $\lambda_{\text{ece}}$ plays an important role in tuning the ECE. Five models were trained with progressively increasing $\lambda_{\text{ece}}$ values (0.0--1.0) to evaluate the effect of keypoint calibration on uncertainty, using a subset (n = 5) of the Shriners data for calibration before applying the optimal setting to the full dataset.  Post-hoc analysis of confidence interval coverage identified $\lambda_{\text{ece}} = 0.5$ as the optimal setting, with a fairly wide range where calibration was not highly sensitive. Thus, we used $0.5$ for the other 63 participants. For the kinematic data, $\lambda_{\text{ece}}$ followed a linear annealing schedule: it remained zero for the first half of training, causing the system to collapse to narrow, overly-confident estimates, and is then increased linearly to its full value by the final iteration to increase the models uncertainty. The processing time varies by system hardware and number of trials, on average a single session, 5 trials, was processed in \~ 50 minutes on a 48GB GPU.

\subsubsection{Expected Calibration Error}
We followed the framework developed in \cite{pierzchlewicz_multi-hypothesis_2022} to provide a global measurement of the calibration (ECE) of the probabilistic model against a ground truth for both spatial and kinematic measurements. At a high level, the ECE models the predicted distribution of errors. It then measures if the empirical distribution of errors match these predictions. This is done by using the Probability Integral Transform (PIT) to transform the empirical errors through the CDF into what should be a uniform distribution, provided the model is calibrated.  A probability-probability (P-P) plot is used to measure the goodness of fit to a uniform distribution\cite{gneiting_probabilistic_2007}. Elaborating further, for each error we compute $u_i=CDF_i(e_i)$ and a calibrated model will then have $u_i \sim \mathrm{Uniform}(0, 1)$. We computed  the ECE both for spatial metrics (step and stride length) and kinematic errors. For spatial metrics, we first compute the predicted empirical distribution of errors defined as  $E = \lVert y_s - \tilde{x} \rVert$ where $y_s$ are the sampled metric  values and $ \tilde{x} $ is the sample median.  Specifically, 1000 samples during the time point of median heel position during each stance phase were passed through the forward kinematic model obtaining sampled heel positions that were then used for step and stride length calculations. Then, given a ground truth of the step metric $x_{gt}$ we compute the measured absolute error as ~$e = \lVert x_{gt} - \tilde{x} \rVert$. Given the empirical error distribution (computed from $S$ sampled step metric values, we can then obtain the empirical CDF value of the measured error defined by $u$, finding where the measured error $e$ sits in the empirical distribution $E$ as $u = \frac{1}{S} \sum_{i=1}^S \mathbf{1}\!\left(E_i \le e\right)$ (i.e., the quantile error each individual step). The $u$ are then aggregated across all the  $N$ steps,  sorted, and compared against the uniform distribution. We then define the expected uniform quantiles as $p_i = \frac{i}{N},\quad i = 1, \dots,N$ and calculate the ECE as ~$\mathrm{ECE}= \frac{1}{N}\sum_{i = 1}^N \big| u_{(i)} - p_i \big|$.   For kinematics, a similar method to calculate the ECE was used. However, since the joint angle probabilities are modeled as a Gaussian distribution, there is no need for an empirical distribution and rank statistics. Instead, each time point, $e_t$, was mapped through the $\mathrm{HalfNormal}$ CDF, with the model predicted scale $\sigma_t$ to obtain transformed values ${u}_i$.

\subsubsection{Spatial Errors and GaitRite Validation}
The spatial calibration was determined by comparing the step length and stride length obtained from the GaitRite instrumented walkway, a validated method for measuring spatiotemporal gait parameters \cite{bilney2003concurrent, mcdonough2001validity}.  GaitRite defines step length as the distance lengthwise along the walkway from the heel center of the current footprint to the heel center of the previous footprint of the opposite foot. Stride length is defined as the distance between the heel position of two consecutive footprints of the same foot. To match these definitions for our probabilistic model, we tracked the calcaneus marker position during the GaitRite-defined contact and toe-off times and defined the heel center time point as the median time point of the stance phase of each foot. The markerless motion coordinate frame was transformed to align with the lengthwise direction of the GaitRite as in \cite{cotton_improved_2023}. To determine the uncertainty of the the model in terms of spatial error we take 1000 samples during the heel center time point and pass this through the forward kinematic model to get the sampled heel positions. We take the difference between two consecutive heel points between the opposite feet to get the step length, and the difference between two consecutive heel points of the same foot to get stride length. This gives us a distribution of step length and stride length values for every step. The ECE was then computed as described above and the accuracy was assessed by computing the absolute step length error.

\subsubsection{Joint Angle Errors and Validation}
The marker-based and probabilistic model's systems utilize fundamentally different kinematic models and anatomical coordinate system definitions, which introduce systematic errors when comparing their output. Before performing our analysis, we removed these systematic biases from the probabilistic model data to ensure a valid comparison \cite{antognini_reframe_2025}. The systematic bias for each joint angle was calculated as the mean difference between the marker-based and probabilistic prediction, averaged over each trial. The final bias values, averaged per participant, were then subtracted from the probabilistic model's predictions to establish a corrected reference for comparative validation against the marker-based data. Joint-angle uncertainty was predictive from per-frame standard deviations produced by the probabilistic model and compared with marker-based joint angles.  Joint angle errors were calculated as the difference between the probabilistic model and the marker-based kinematic trajectories across the whole trial. In the table presenting results from the kinematic analysis, data (hip, knee, ankle) were combined across the left and right limbs. Pelvis angles were treated as non-lateralized.

\section{Results}
\subsection{Spatiotemporal Gait Metrics Calibration and Accuracy}
\begin{figure}[!htbp]
    \centering
    \includegraphics[width=1\linewidth]{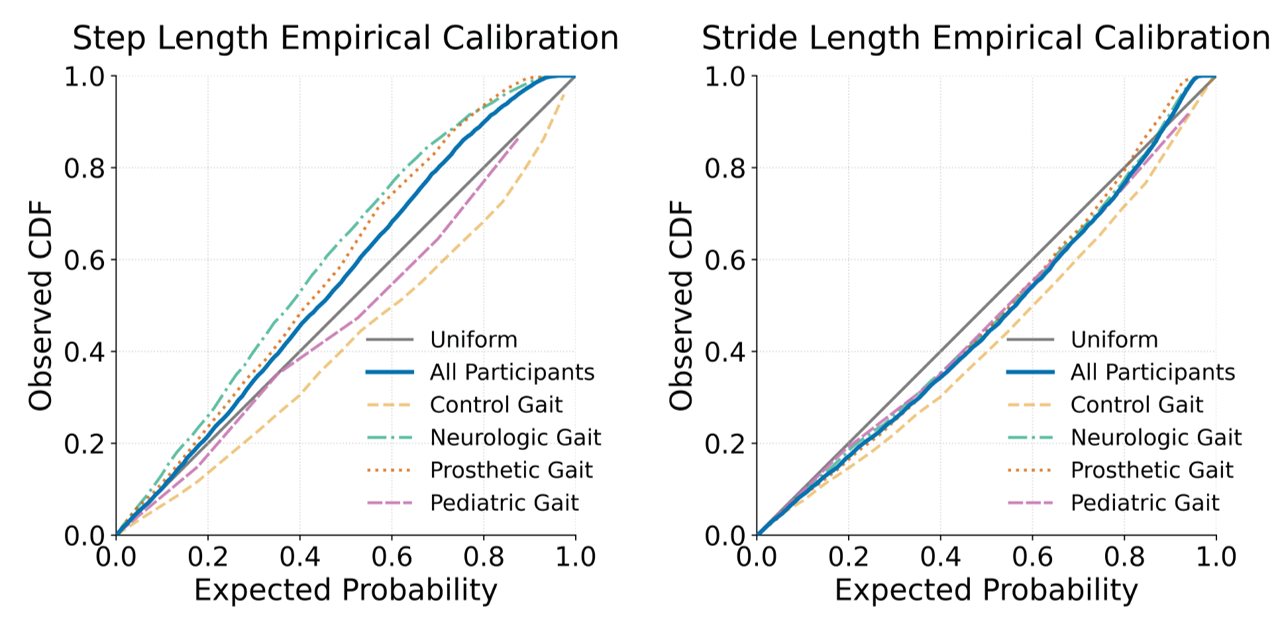}
    \caption{Empirical calibration curve for step length and stride length measures across all participants (blue) and separated by participant groups in different colors. PIT values (dots) are obtained from the posterior predictive absolute error distribution and are plotted against the expected uniform distribution (dashed line). A perfectly calibrated model will have the PIT values fall on the identity line. Deviations from the identity line are indications of miscalculation quantified by ECE (See Table~\ref{tab:step_length_ece_table}).}
    \label{fig:ece_curves}
\end{figure}

Spatial calibration of the model was evaluated by computing the empirical ECE as described previously, and was found to be well calibrated across conditions and populations with $ECE \leq 0.1$, see Table~\ref{tab:step_length_ece_table}. When grouping all the trials based on walking speed, we also found that the model remains consistently calibrated across all speeds $ECE \leq 0.1$.  Visualization of these uncertainty measures in Fig.~\ref{fig:ece_curves}, we see that the model tends to show slightly better calibration overall when its estimation of stride length is compared to the ground truth, rather than step length.

\begin{table}[!htbp]
\centering
\begin{tabular*}{\columnwidth}{@{\extracolsep{\fill}}lcc}
\toprule
\textbf{Participant Group} & \textbf{Step Length} & \textbf{Stride Length} \\
\midrule
All & 0.05 & 0.04 \\
Able-Bodied Controls & 0.08 & 0.07 \\
Neurologic Gait & 0.10 & 0.03 \\
Pediatric Gait & 0.03 & 0.03 \\
Prosthetic Gait & 0.08 & 0.04 \\
\midrule
\textbf{Gait Speed (All Participants)} & & \\
\midrule
Slow & 0.02 & 0.02 \\
Self-Selected & 0.02 & 0.02 \\
Fast & 0.07 & 0.04 \\
\bottomrule
\end{tabular*}
\caption{ECE values for posterior step-length and stride-length estimates.}
\label{tab:step_length_ece_table}
\end{table}

\begin{figure}[!htbp]
    \centering
    \includegraphics[width=0.6\linewidth]{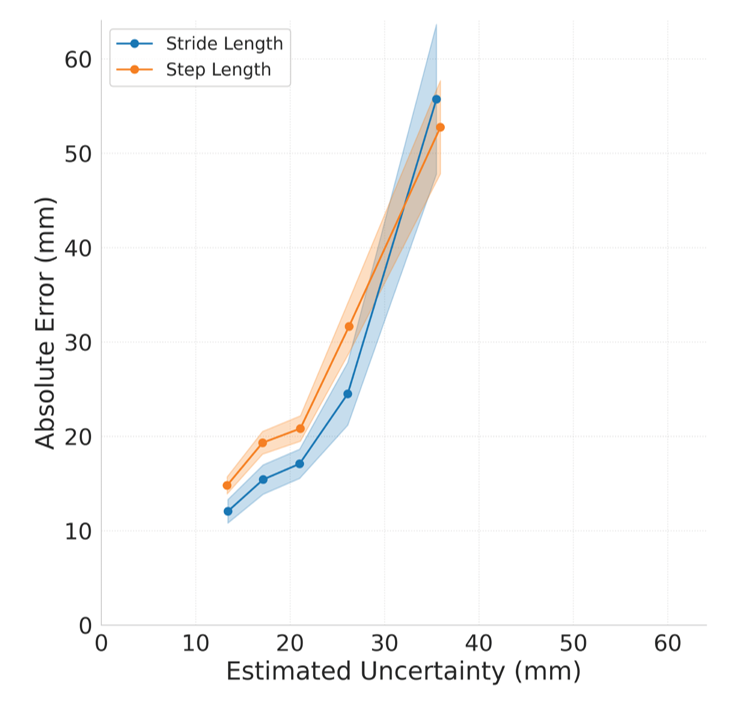}
    \caption{Step length and stride length mean absolute error versus the predicted uncertainty across all participants. The points show the mean absolute error for samples that belong in that uncertainty bin. Shaded regions show the 95\% confidence intervals for the mean absolute error in each uncertainty bin.}
    \label{fig:error_uncertainty_stepStrideLength}
\end{figure}

We also measure the accuracy of our model compared to the instrumented walkway by calculating the difference between the step and stride lengths computed from their median heel positions to the instrumented walkway levels (Tables \ref{tab:step_length_distribution} and \ref{tab:stride_length_distribution}). Fig.~\ref{fig:error_uncertainty_stepStrideLength} shows how this error increases as we bin steps by their increasing uncertainty, consistent with the confidence intervals from the model being calibrated and the uncertainty indicating how much the data should be trusted.

We then evaluated the distribution of predicted uncertainty and the corresponding absolute error for both step and stride length of each group. Results are summarized in  \ref{tab:step_length_distribution} and \ref{tab:stride_length_distribution} where we show the median and interquartile range (IQR) of the uncertainty and absolute error for the entire dataset and for the measurements with the lowest 50\% uncertainty as well as the extrema. By selecting the data with the lowest 50\% uncertainty, the data that the model is more confident in, we obtain consistently lower errors and narrower inter-quartile ranges. This shows that when the model is confident, its measurements are more consistent and accurate. This is very prominent in the neurologic gait group. In our dataset, this is the population most likely to therapists standing nearby for assistance. Overall, they have the highest absolute error and predictive uncertainty. By filtering out steps and strides whose predicted uncertainty exceeded the  $50^{th}$  percentile, we found an overall median step length error of $12.01$ mm and median stride length error of 9.13 mm. We observed the highest $10^{th}$ percentile predictive uncertainty across all participants contains the outliers with high errors and variability in median step length error ($38.73$ mm) and median stride length error ($30.82$ mm). When the model's uncertainty is small, we obtain very accurate results with reduced variability ($9.85$ mm for step length, and $7.31$ mm for step width error). Notably, these uncertainty estimates come from the model and do not require the use of external ground truth to determine.

\begin{table}[!htbp]
\centering
\begin{tabular}{lcccc}
\toprule
\textbf{Step Length} & \multicolumn{2}{c}{\textbf{Uncertainty (mm)}} & \multicolumn{2}{c}{\textbf{Error (mm)}} \\
& Median & IQR & Median & IQR \\
\midrule
All Participants & 21.09 & 11.68 & 16.24 & 25.60 \\
Controls         & 17.55 &  8.30 &  9.08 & 12.29 \\
Neurologic       & 24.68 & 12.91 & 23.14 & 31.39 \\
Prosthetic       & 18.44 &  8.35 & 16.07 & 23.44 \\
Pediatric        & 17.28 &  8.32 & 10.68 & 12.21 \\
\midrule
\textbf{$<50^{th}$ Percentile} & \multicolumn{2}{c}{} & \multicolumn{2}{c}{} \\
\midrule
All Participants & 16.18 & 4.60 & 12.01 & 17.59 \\
Controls         & 13.93 & 3.66 &  7.66 &  9.04 \\
Neurologic       & 18.68 & 6.17 & 17.33 & 24.44 \\
Prosthetic       & 14.97 & 3.43 & 15.73 & 22.42 \\
Pediatric        & 13.67 & 2.65 & 11.16 & 11.03 \\
\midrule
\textbf{Extrema Uncertainty} & \multicolumn{2}{c}{} & \multicolumn{2}{c}{} \\
\midrule
Noisiest $10^{th}$ Percentile  & 42.96 & 11.82 & 38.73 & 55.45 \\
Accurate $10^{th}$ Percentile  & 11.94 &  1.69 &  9.85 & 12.31 \\
\bottomrule
\end{tabular}
\caption{Step-length uncertainty estimates from the probabilistic model and corresponding absolute errors relative to the GaitRite. Results are reported for all participants and for each subset. Uncertainty is further stratified into the most confident trials ($<50^{\text{th}}$ percentile) for each subset and the most uncertain trials (top $10^{\text{th}}$ percentile) and the least uncertain trials (bottom $10^{\text{th}}$ percentile) across all participants.}
\label{tab:step_length_distribution}
\end{table}

\begin{table}[!htbp]
\centering
\begin{tabular}{lcccc}
\toprule
\textbf{Stride Length} & \multicolumn{2}{c}{\textbf{Uncertainty (mm)}} & \multicolumn{2}{c}{\textbf{Error (mm)}} \\
& Median & IQR & Median & IQR \\
\midrule
All Participants & 21.01 & 11.41 & 11.83 & 17.65 \\
Controls         & 17.45 &  7.73 &  8.86 & 13.04 \\
Neurologic       & 24.59 & 12.67 & 13.90 & 20.89 \\
Prosthetic       & 18.04 &  7.93 & 10.62 & 17.60 \\
Pediatric        & 17.17 &  7.91 & 10.87 & 12.80 \\
\midrule
\textbf{$<50^{th}$ Percentile} & \multicolumn{2}{c}{} & \multicolumn{2}{c}{} \\
\midrule
All Participants & 16.20 & 4.43 &  9.13 & 12.56 \\
Controls         & 14.11 & 3.39 &  7.30 &  8.95 \\
Neurologic       & 18.74 & 6.06 & 10.39 & 13.90 \\
Prosthetic       & 15.01 & 3.19 &  9.23 & 14.06 \\
Pediatric        & 13.71 & 3.33 &  9.89 & 11.76 \\
\midrule
\textbf{Extrema Uncertainty} & \multicolumn{2}{c}{} & \multicolumn{2}{c}{} \\
\midrule
Top $10^{th}$ Percentile     & 43.27 & 11.78 & 30.82 & 47.94 \\
Bottom $10^{th}$ Percentile  & 12.15 &  1.55 &  7.31 &  8.47 \\
\bottomrule
\end{tabular}
\caption{Stride-length uncertainty estimates from the probabilistic model and corresponding absolute errors relative to the GaitRite. Results are reported for all participants and for each subset. Uncertainty is further stratified into the most confident trials ($<50^{\text{th}}$ percentile) for each subset and the most uncertain trials (top $10^{\text{th}}$ percentile) and the least uncertain trials (bottom $10^{\text{th}}$ percentile) across all participants.}
\label{tab:stride_length_distribution}
\end{table}

\subsection{Systematic Biases and Trajectory Agreement}
The marker-based and MMMC systems utilized fundamentally different kinematic models and anatomical coordinate system definitions, which necessitated the calculation and correction of systematic biases for comparative analysis. As detailed in Table \ref{tab:Joint_Biases_Summary}, on average, large mean differences were observed in joints such as pelvis tilt ($19.71^{\circ} \pm 4.91^{\circ}$) and hip flexion ($23.25^{\circ} \pm 6.36^{\circ}$), while pelvis obliquity showed a near-zero bias ($-0.22^{\circ} \pm 2.75^{\circ}$). These constant, participant-specific biases were calculated and applied as a correction to the probabilistic reconstructions to ensure that subsequent analyses reflected tracking fidelity rather than modeling discrepancies.

\begin{figure}[!htbp]
\centering
\includegraphics[width=1\linewidth]{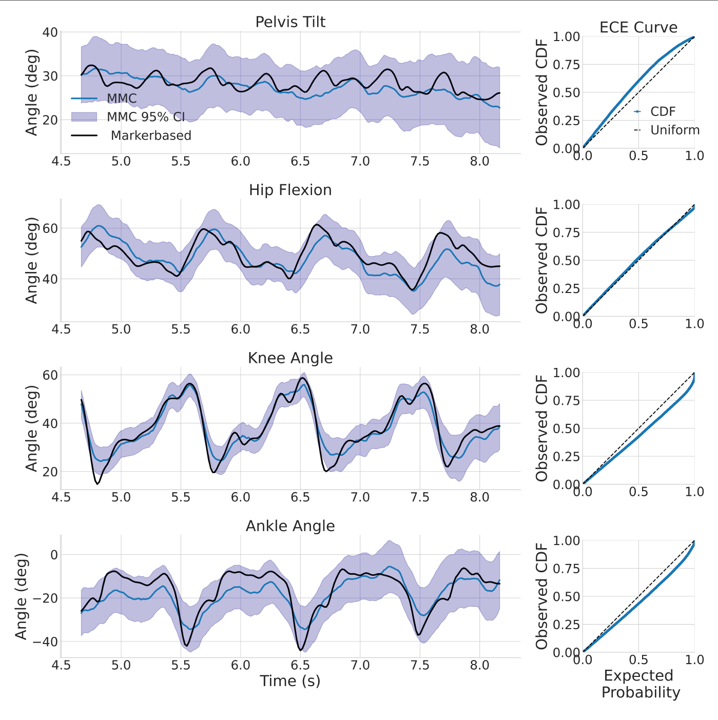}
\caption{Sagittal-plane joint kinematics and uncertainty calibration results for pelvis tilt, hip flexion, knee flexion, and ankle dorsiflexion. Left column: Probabilistic model's predictions are shown as the mean trajectory (solid blue) with 95\% confidence intervals (shaded region). Marker-based kinematics are shown for reference (dashed gray). Right column: Empirical values from the PIT, CDF and the dashed diagonal line indicates perfect calibration.}
\label{fig:Joint_Angle_Composite}
\end{figure}

Following bias correction, the probabilistic model's time-series output closely matched the marker-based trajectories across the lower extremity joints. As illustrated by the mean and $95\%$ confidence interval in the first column of Fig. \ref{fig:Joint_Angle_Composite}, the probabilistic model's output successfully captured general kinematic patterns but did not exactly match the marker-based dynamics at the pelvis which was then propagated to the hip joint.

\begin{table}[!htbp]
    \centering
    \begin{tabular}{l r @{\hspace{3pt}} c @{\hspace{3pt}} l}
    \toprule
    \textbf{Joint} & \multicolumn{3}{c}{\textbf{Mean $\pm$ SD (deg)}} \\
    \midrule
    Pelvis Tilt       & 19.71   & $\pm$ & 4.91  \\
    Pelvis Obliquity  & $-0.22$ & $\pm$ & 2.75  \\
    Pelvis Rotation   & 0.76    & $\pm$ & 4.27  \\
    Hip Flexion       & 23.25   & $\pm$ & 6.36  \\
    Hip Abduction     & 4.56    & $\pm$ & 1.84  \\
    Hip Rotation      & 2.19    & $\pm$ & 10.29 \\
    Knee Flexion      & 3.19    & $\pm$ & 4.60  \\
    Ankle Flexion     & 2.73    & $\pm$ & 6.46  \\
    \bottomrule
    \end{tabular}
\caption{Biases between marker-based and probabilistic reconstructions at each lower limb joint.}
\label{tab:Joint_Biases_Summary}
\end{table}

\subsection{Kinematic Calibration}
\begin{figure*}[!tbp]
\centering
\includegraphics[width=1\linewidth]{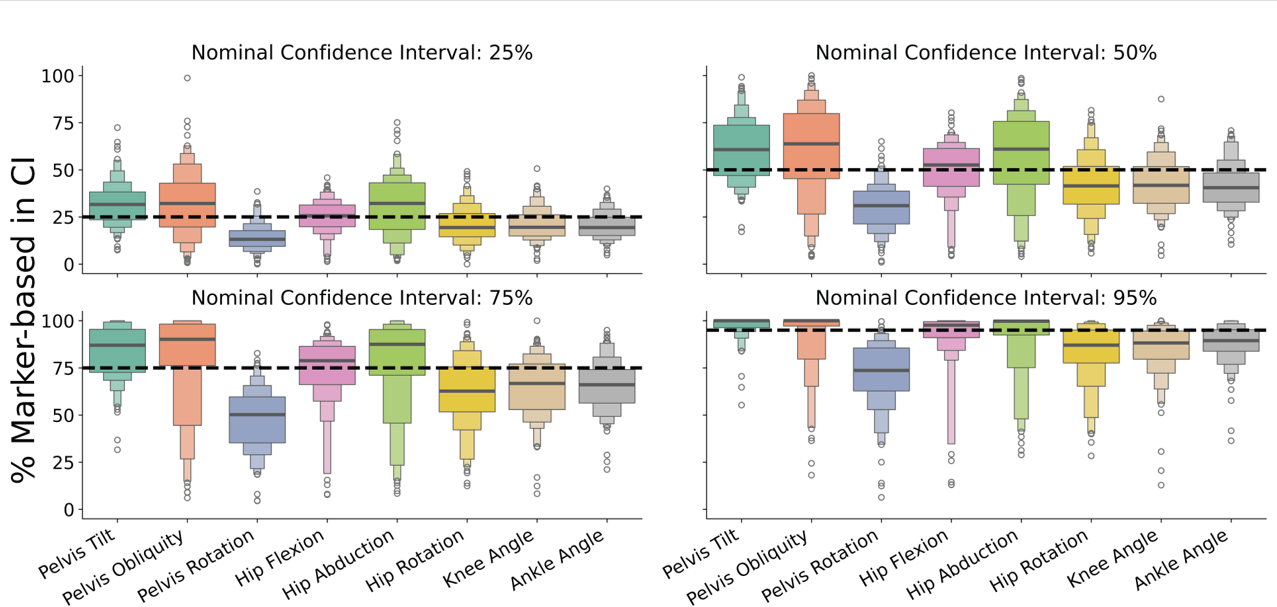}
\caption{Percentage of marker-based joint angles that fall within the probabilistic model’s nominal confidence intervals (25\%, 50\%, 75\%, and 95\%) across eight lower-extremity kinematic variables, on a per-trial basis. Each panel corresponds to a different nominal confidence level, with the dashed horizontal line indicating ideal coverage. This figure summarizes how well the model’s predicted uncertainty aligns with ground-truth marker-based kinematics across joints and confidence levels.}
\label{fig:nominal_joint_angle_coverage}
\end{figure*}

For each joint, we computed the ECE to quantify the calibration of the uncertainty outputs from the probabilistic reconstructions. The ECE values demonstrated  the probabilistic model was generally well-calibrated across lower extremity joints (ECE $<0.10$), with the notable exception of pelvis rotation, as shown in Table \ref{tab:joint_noise_error}. Further assessment of the median percentage of time the marker based values were within the nominal confidence intervals of the model are shown in Fig. \ref{fig:nominal_joint_angle_coverage}. If the predicted uncertainty of the model was too large, then the proportion of the marker-based data in within the confidence interval would be above the nominal line.  Conversely, if the predicted uncertainty from the model was too small, then the proportion of the marker based data would be beneath than the nominal line. For example the ECE value for hip flexion is $0.01$ with very small deviations in the percentage of the time the marker data falls within the nominal confidence interval showing great calibration. While with pelvis rotation, which has an ECE of ($0.18$), the median \% time the marker based data is within the nominal confidence interval is much lower Fig. \ref{fig:nominal_joint_angle_coverage}.

The distribution of both the predicted uncertainty and the errors between the systems exhibited clear joint-dependent differences, as detailed in Table~\ref{tab:joint_noise_error}. Generally, the magnitude and spread of the predicted errors closely tracked the probabilistic model's predicted uncertainty. Across all joints, the median predicted uncertainty remained below $4.78^\circ$, and the median absolute error remained below $3.76^\circ$.   The joints with the lowest predicted uncertainty: pelvis tilt and pelvis obliquity (median uncertainty $\approx 3.28^\circ$) also exhibited the lowest median errors ($\approx 1.52$--$1.62^\circ$). Conversely, the joints with the largest predicted uncertainty: hip rotation, knee flexion, and ankle flexion (median uncertainty $\approx 3.56$--$4.78^\circ$) demonstrated the largest median errors ($\approx 2.84$--$3.76^\circ$). The exception to this was pelvis rotation, where the predicted uncertainty (median $1.99^\circ$) was the lowest of all joints, yet the corresponding median absolute error ($2.15^\circ$) was significantly higher than the uncertainty, a mismatch consistent with the greatest miscalibration (ECE of $0.18$).

\begin{table}[!htbp]
\centering
\begin{tabular*}{\columnwidth}{@{\extracolsep{\fill}}lcccccc}
\toprule
\textbf{Joint} & \textbf{ECE} & \multicolumn{2}{c}{\textbf{Uncertainty ($^\circ$)}} & \multicolumn{2}{c}{\textbf{Error ($^\circ$)}} \\
\cmidrule(lr){3-4} \cmidrule(lr){5-6}
 &  & Median & IQR & Median & IQR \\
\midrule
Pelvis Tilt        & 0.07 & 3.28 & 0.76 & 1.62 & 0.81 \\
Pelvis Obliquity   & 0.06 & 3.23 & 0.67 & 1.52 & 1.27 \\
Pelvis Rotation    & 0.18 & 1.99 & 0.33 & 2.15 & 1.34 \\
Hip Flexion        & 0.01 & 4.26 & 0.99 & 2.53 & 1.13 \\
Hip Abduction      & 0.05 & 3.74 & 0.87 & 1.88 & 1.55 \\
Hip Rotation       & 0.08 & 4.78 & 1.20 & 3.76 & 1.55 \\
Knee Flexion       & 0.06 & 3.56 & 0.90 & 2.84 & 1.22 \\
Ankle Flexion      & 0.07 & 4.73 & 1.31 & 3.52 & 1.04 \\
\bottomrule
\end{tabular*}
\caption{ECE, predicted uncertainty and absolute errors for each kinematic trajectory with median and IQR across each joint.}
\label{tab:joint_noise_error}
\end{table}

\section{Discussion}

The overall purpose of this study was to validate whether the probabilistic reconstructions were biomechanically accurate and output well-calibrated uncertainty estimates. {Which entails biomechanical accuracies lower than either the MDC or the MCID, and well calibrated uncertainty estimates in which a nominal predicted (X\%) confidence interval contains close to X\% of the clinically measured data.

{We found that the probabilistic model accurately tracks body segment positions and joint angles and provides a well-calibrated uncertainty metric. The step length and stride length, as measured by the instrumented walkway, are both accurate and well calibrated across all participants. Correspondingly, trials with greater estimated uncertainty demonstrated greater error against ground truth. We also found that the kinematic patterns across lower extremity joints, other than pelvic obliquity, are accurate and well calibrated up to a participant based offset.

Importantly, our findings indicate the probabilistic method indicates when we can trust our computer-vision based probabilistic model. This unlocks the potential for removing the reliance on additional, external clinical systems to provide ground truth measures for validation, and providing a quality-control tool for the MMC output to produce trusthworhty results.

\subsection{Spatiotemporal Gait Metrics}

For spatiotemporal metrics, the ECE over all participants for step length was $0.05$ and stride length was $0.04$.  The median step length error was $16.24$ mm and median stride length error was  $11.83$ mm, which is below the MDC of 12.0 $\pm$ 5.4  \cite{paratietal_2022}  This also demonstrates internal consistency between the the uncertainty estimates in the current work and what was shown previously for spatial errors with virtual marker locations \cite{cotton_confidence_2025}.  Overall, these results confirm that the confidence intervals from the probabilistic reconstructions are statistically calibrated for spatial gait parameters when compared to an instrumented walkway.

Fig. \ref{fig:error_uncertainty_stepStrideLength}, which shows how absolute error varies across the predicted uncertainty over all the participants, showed the lowest mean absolute error of around $\sim 13$ mm for both step and stride length. The error can in part be attributed to the differences between the GaitRite identified pressure points and the probabilistic pose estimation markers. The larger errors seen in the neurologic and prosthetic group could be due to different loading patterns during initial contact in these individuals. This could also explain the slightly higher step length error compared to the stride length error.

Importantly, Tables \ref{tab:step_length_distribution} and \ref{tab:stride_length_distribution} show that we can use the predicted uncertainty to filter out noisy values and trust that we are left with high-quality data. As seen, choosing steps where the predicted uncertainty was below the 50\textsuperscript{th} percentile led to greatly reduced absolute error medians as well as a lower spread in the error.  Critically, we can now use the model's uncertainties to filter out trials with high error, without reliance on validation from an instrumented walkway, which is a crucial component for deploying this system outside of the gait lab.

\subsection{Kinematics}
Previous MMMC studies have consistently reported near-constant biases between marker-based and markerless kinematics, particularly for pelvic orientation \cite{antognini_reframe_2025}. These discrepancies arise from differences in model definitions—such as joint center locations, segment lengths, and other anthropometric constraints—that are explicitly embedded in marker-based reconstructions but optimized as scaling parameters in the probabilistic model outputs. We assessed the kinematic accuracy of the lower-extremity joint angles across the entire respective waveform. We observed the median, bias corrected error between systems to be <$3.76 ^\circ$,  within the MCID of < $5^{\circ}$ (Table~\ref{tab:joint_noise_error}) \cite{mcginley_reliability_2009}}. This provides evidence that the probabilistic models are comparable to marker-based motion capture up to an offset which is consistent with previous MMMC research \cite{antognini_reframe_2025, wren_theia_2023}.

The confidence intervals output from the probabilistic model are shown to be fairly well calibrated across each of the joints, similar to \cite{cotton_confidence_2025}.  The small ECE values indicate the estimated uncertainty is well-calibrated to the measured error between the systems, (Table~\ref{tab:joint_noise_error}). Overall, the probabilistic model’s predicted noise closely reflected the actual errors between the marker-based and probabilistic reconstruction. Pelvis rotation notably deviates from this trend, being substantially overconfident with an ECE of 0.18 and only $\sim$70\% of the samples falling within the 95\% confidence interval Fig. \ref{fig:nominal_joint_angle_coverage}. We suspect this discrepancy stems from the order the Euler angles are applied to the biomechanical models to define pelvis orientation. In the MuJoCo model, rotation is applied last, whereas in the SCGM model, pelvis rotation is applied first. While this is the most problematic source of mis-calibration, other structural differences in our MuJoCo model limit a perfect mapping to clinical-grade motion analysis, which led us to apply the per-joint bias corrections. Refining the MuJoCo model for MMMC is an important future direction to improve the clinical utility of our probabilistic MMMC approach.

\subsection{Limitations}
One of the primary limitations of this study was that all of the data was collected in a gait laboratory setting. The environment was a large, open space with a designated 10m walkway for the patient and therapist, and had optimal, unobstructed camera coverage.  In practice, these MMMC systems will be deployed in busy clinical spaces, with no clearly defined walking paths, multiple patients, therapists, or caretakers in the collection space, and potentially suboptimal camera placements, due to clinical or privacy constraints. Notably, we used this probabilistic algorithm on data collected from MMMC systems deployed in therapy gyms, and initial tests suggest the predicted uncertainty was successful in filtering low-quality data caused by transient occlusions. With the kinematic data collected in the gait laboratory with eight camera views, and time-synchronized with the marker-based motion capture collected in the middle of the collection volume, truly limits our ability to test the use of uncertainty estimates for the exclusion of noisy or unreliable data.  Future work will expand on camera ablations, and how that effects the changes in the estimated uncertainty, similar to \cite{cotton_confidence_2025}.

The per-participant bias removal is another limitation inherent to the current analysis. As noted above, optimizing the biomechanical model used in our probabilistic MMMC to be more aligned to clinical models is an important future direction that we hope will both improve calibration while reducing these biases.

The confidence intervals from the probabilistic model provide a robust method to identify where the model is uncertain and the ability to exclude unreliable data. This also raises important questions about what causes this increased uncertainty. Anecdotally, we have seen that occlusions from other people, assistive devices, and limb characteristics that differ from those of able-bodied individuals are among the factors that contribute to model uncertainty. In future work, we plan to analyze this more quantitatively using the confidence estimates from this probabilistic method to systematically evaluate the reason for higher uncertainty between individual trials and across different clinical populations. We will also optimize the model training and hyperparameters with a unified annealing schedule. We will further assess other statistical metrics beyond ECE (global probabilistic calibration) with a unified annealing schedule, larger datasets, and to assess the effects of temporal errors with conditional probabilities e.g. errors as a function of percent gait cycle or errors with respect to joint angle magnitude or joint angle velocity dependent errors, across difference clinical populations.

\section{Conclusion}
We validated a MMMC-based probabilistic method, that reconstructs kinematics and outputs well-calibrated confidence intervals, against external validation data -- spatial gait parameters from an instrumented walkway and kinematics from marker-based motion analysis -- across a diverse clinical population. While outputs from MMMC systems need to be accurate, it is also essential to know \emph{when} they can be trusted to exclude unreliable inferences. This MMMC-based probabilistic framework is an important step beyond traditional validation studies that speak to the aggregate model performance, but cannot identify the outliers and unreliable inferences. This validation study shows that our probabilistic method can provide calibrated confidence intervals at the level of each joint and instance for a participant, allowing for a trustworthy gait analysis pipeline, which is essential for MMMC to be used in wider practice.

\section*{Acknowledgments}
The contents of this paper were developed under a grant from the National Institute on Disability, Independent Living, and Rehabilitation Research (NIDILRR grant number 90REGE0030) and with support from the National Institute of Child Health and Human Development and the National Center for Medical Rehabilitation Research (R01HD114776) (both RJC). FS is supported by the European Research Council (ERC) under the European Union's Horizon Europe research and innovation program (Grant agreement No. 101171526).

{\small
\printbibliography
}

\appendix
\section{Hyperparameters and Network Settings}
\label{sec:appendix}

Briefly, the hyperparameter of the probabilistic model we used, based on our work in \cite{cotton_differentiable_2024}, we use a GPU-accelerated and differentiable biomechanical model running in MuJoCo \cite{todorov_mujoco_2012}. The output trajectories are used to train an implicit representation $p_{\phi}(\theta_{t})$ as an MLP $f_{\phi}$ that takes time as input and outputs the mean and covariance of the joint angle distribution. The specific hyperparameter, configurations and network settings used for the probabilistic model can be found in Table \ref{tab:model_settings}. More details, including implementation specifications can be found in \cite{cotton_confidence_2025}.

\begin{table*}[!htbp]
\caption{Hyperparameter Settings for Probabilistic Reconstruction}
\label{tab:model_settings}
\begin{tabularx}{\textwidth}{@{} p{4cm} p{3cm} X @{}}
\toprule
\textbf{Parameter} & \textbf{Value} & \textbf{Description} \\
\midrule
Maximum Iterations & 30000 &  Total iterations. \\
Layer Normalization & false & Normalize weights of the MLP layers. \\
Reproject Keypoint Noise & true & Reproject noise into the image space. \\
ECE Keypoint Percentage & 25.0 & Percentage of keypoints used in the ECE calculation. \\
Keypoint Distribution Scalar & true & Use a scalar distribution for keypoint uncertainty. \\
Keypoint Distribution Type & HalfLaplace & Distribution of keypoint uncertainty. \\
Trajectory Sample Length & 100 & Length of trajectory samples (frames). \\
Initial Standard Deviation & [0, 1, -20] & Initial standard deviation for trajectory components. \\
Mean Sample NLL Weight & 0.0 & Weight of the NLL of mean samples in the loss. \\
Reprojection Loss Weight & 0.0 & Weight of reprojection error during optimization. \\
Layer Dimensions & [128, 256, 512, 1024] & Hidden layer dimensions in the MLP. \\
Covariance Distribution Rank & 20 & Rank of the low-rank covariance approximation. \\
Site Offset Regularization Strength & $10^{4}$ & Regularization for body site offset parameters. \\
Initial Learning Rate & $10^{-3}$ & Initial learning rate of the optimization. \\
ELBO Confidence Weight & 0.0 & Weight of the confidence term in the ELBO. \\
ECE Regularization Weight & 0.5 & Weight for the ECE regularization loss. \\
Anneal ECE Loss & false (Spatial), true (Kinematics) & ECE loss weight is annealed during training. \\
ECE Maximum Scale & 1.0 & Maximum scaling factor for ECE annealing. \\
Variational Samples per Gradient Step & 8 & Number of samples from the variational posterior used to estimate the ELBO. \\
Keypoint Likelihood Smoothing & 200 & Smoothing parameter for keypoint likelihood. \\
HalfLaplace Half-Width & 30 & Half-width parameter of the HalfLaplace distribution. \\
Temporal Smoothness Regularization & 10 & Regularization weight for temporal smoothing of the kinematic trajectory. \\
Robust Multi-Camera Weighting & true & Whether to use dynamic robust weighting across camera observations. \\
Camera Confidence Masking Threshold & 0.7 & Confidence threshold for masking unreliable cameras. \\
\bottomrule
\end{tabularx}
\end{table*}

\end{document}